\documentclass[runningheads]{llncs}
\usepackage[T1]{fontenc}
\usepackage{graphicx}
\usepackage{array}
\usepackage{booktabs} 
\usepackage{longtable}

\begin{document}
%

%==================================
% Option 1
% \title{Pedagogy-Driven Evaluation of AIED Tutoring Systems Grounded in Learning Science Principles}

%option 2
\title{Pedagogy-driven Evaluation of Generative AI-powered Intelligent Tutoring Systems}

% Option 3
% \title{Pedagogy-Driven Evaluation of Generative AI-powered Tutoring Systems Grounded in Learning Science Principles}
%==================================o

\titlerunning{Pedagogy-driven Evaluation of ITS}
\author{Kaushal Kumar Maurya\orcidID{0000-0003-0249-6508}\\ \and
Ekaterina Kochmar\orcidID{0000-0003-3328-1374}}
\authorrunning{Maurya and Kochmar}

\institute{EduNLP Lab, Mohamed bin Zayed University of Artificial Intelligence,\\ Abu Dhabi, UAE \\
\email{\{kaushal.maurya, ekaterina.kochmar\}@mbzuai.ac.ae}}
\maketitle              % typeset the header of the contribution

%\textbf{Expectations:} BlueSky paper submissions do not necessarily require new empirical results, unlike more traditional AIED submissions. Despite BlueSky submissions’ focus on novel, exploratory solutions for the future, there is still the need to support their ideas with sufficient evidence. When BlueSky submissions focus on novel perspectives on existing problems or a new research vision, as examples, they might not require empirical results. However, such submissions are still expected to defend their positions via robust scientific argument rooted in the relevant literature. A thorough exploration of implications, with detailed discussions, are considered important as well.

\vspace{-0.2cm}
\begin{abstract}
The interdisciplinary research domain of Artificial Intelligence in Education (AIED) has a long history of developing Intelligent Tutoring Systems (ITSs) by integrating insights from technological advancements, educational theories, and cognitive psychology. The remarkable success of generative AI (GenAI) models has accelerated the development of large language model (LLM)-powered ITSs, which have potential to imitate human-like, pedagogically rich, and cognitively demanding tutoring. However, the progress and impact of these systems remain largely untraceable due to the absence of reliable, universally accepted, and pedagogy-driven evaluation frameworks and benchmarks. Most existing educational dialogue-based ITS evaluations rely on subjective protocols and non-standardized benchmarks, leading to inconsistencies and limited generalizability. In this work, we take a step back from mainstream ITS development and provide comprehensive state-of-the-art evaluation practices, highlighting associated challenges through real-world case studies from careful and caring AIED research. Finally, building on insights from previous interdisciplinary AIED research, we propose three practical, feasible, and theoretically grounded research directions, rooted in learning science principles and aimed at establishing fair, unified, and scalable evaluation methodologies for ITSs.

\keywords{Intelligent Tutoring Systems (ITSs)  \and Generative AI (GenAI) \and Large Language Models (LLMs) \and Evaluation \and Pedagogy}
\end{abstract}

\section{Introduction}
\vspace{-0.2cm}
% Fundamental details about the progress of generative AI, opportunities to fine-tune LLMs as AI tutors, brief details about associated challenges/gaps in evaluation, and the path forward.
% Add a flow figure for this paper.

% Paragraph-01: General introduction about LLM, Educational spce and ITS/AI Tutors 
Over the past decades, Artificial Intelligence (AI) has undergone multiple paradigm shifts, each bringing new opportunities and challenges. However, the emergence of large language models (LLMs) in the current era of generative AI (GenAI) marks the most notable breakthrough \cite{zhao2023survey}, with transformative impact across numerous domains. One of the most pressing global challenges is ensuring high-quality and equitable education -- a cornerstone of societal progress. The educational domain is now actively exploring ways to harness the power of LLMs to bridge this gap, aiming for universal access to personalized and effective learning experiences. AI methods have been part of AIED research for many years \cite{pinkwart2016another}. Recent advances in GenAI have pushed the boundaries of AI-driven educational applications \cite{yan2024practical}, including automated student feedback generation, intelligent tutoring systems (ITS), lesson planning, and more. Among these, ITSs, also referred to as AI Tutors, have garnered significant attention. The rise of conversational LLMs closely aligns with the long-standing vision of ITSs which are interactive, adaptive, and scalable learning environments. This has led to the development of \textit{educational dialogue-based ITSs} like LearnLM \cite{team2024learnlm}, Khanmigo \cite{khanmigo2023}, and similar.

% Paragraph-02: Introduction to the evaluation of ITS, brief history and Challenges in evaluating ITS

Despite the active development of ITSs, their true progress and impact remain difficult to quantify due to the absence of universally accepted pedagogical principles, robust evaluation frameworks, and comprehensive benchmarks \cite{jurenka2024towards}. One of the primary challenges lies in the inherently multidisciplinary nature of educational research, where existing studies in cognitive and learning sciences are often conducted on small, homogeneous populations, primarily from WEIRD (Western, Educated, Industrialized, Rich, and Democratic) countries, with variable implementation parameters \cite{pinkwart2016another}. This has led to the development of subjective, personalized, or limited-scale evaluation frameworks and benchmarks that lack generalizability and broad acceptance \cite{tack-etal-2023-bea,tack2022ai}. For instance, most traditional assessments of ITSs rely on \textit{self-reported} outcomes \cite{goe2008approaches}, which are inadequate for modern, interactive, and on-the-fly GenAI-based ITSs. Additionally, common natural language generation (NLG) metrics are not designed to capture pedagogical values of interactions, often requiring an optimal single reference and being susceptible to adversarial inputs \cite{tack-etal-2023-bea}. Finally, efforts to develop evaluation models that capture pedagogical effectiveness are either incomplete, failing to encompass the full range of pedagogical nuances, limited to subjective and non-scalable human evaluations, or restricted to narrow domains and specific use cases \cite{jurenka2024towards}. Addressing these limitations is crucial for advancing the reliability and scalability of ITS evaluation methodologies.

% Paragraph-03: Case studies which show due to lack of proper evaluation metrics there are already bad outcomes.  

The lack of appropriate evaluation protocols affects both human tutoring and ITSs. In the ``No Child Left Behind Act'' (NCLB) in the United States, overemphasis on standardized test scores led to shallow learning, reduced engagement, and limited teaching methods \cite{au2007high,hargreaves2003teaching,hursh2007assessing}. Similarly, Automated Essay Scoring (AES) systems driven by LLMs focus on surface-level features like grammar and vocabulary, often neglecting critical thinking and conceptual understanding \cite{haswell2006automated,perelman2012construct,perelman2014critiquing}. This limits teachers' insight into students' reasoning and hinders personalized instruction \cite{whitlock2020limitations}. Finally, a field experiment in a Turkish high school involving nearly 1,000 students from grades 9 to 11 examined the impact the best state-of-the-art GenAI model (GPT-4) as tutor on math learning~\cite{Sungu2024GenerativeAI}. Over four 90-minute sessions covering 15\% of the curriculum, students engaged in reviewing lectures, AI-assisted practice, and unassisted exams. While AI-supported students solved 48\% more practice problems correctly, they scored 17\% lower on unassisted tests, suggesting that reliance on AI during practice may have impeded the development of independent problem-solving skills. These cases highlight the need for evaluation metrics that assess conceptual clarity, engagement, pedagogical effectiveness, and critical thinking alongside human-like responses.

% Paragraph-04: Organization of the rest of the paper / key highlights / key contributions
% This work present multifold contribution listed below: 
% \begin{enumerate}
%     \item We present \textit{the current state of evaluation methodologies} for ITSs by providing a comprehensive review of AIED evaluation frameworks and pedagogical principles. This review covers a broad spectrum, ranging from evaluation approaches based on self report for traditional, static/rule-based ITS, to NLG and pegagogical-orgiented evalaution for advance deep-leaning and genAI based ITS.
%     \item We pin point key challenges in evaluation of ITS by highlighting the diverse pragmatic of evaluation and non-unification of the exiting principles, evaluation methodologies and benchmarks. 
%     \item Finally, we proposed three main research directions grounded in learning science principles and previous research in AIED, as step towards migiting the key challegnes whihc aimed to be unified, saclable, and help to track the progress of quality of ITS.
% \end{enumerate}

This work makes several key contributions, listed below:

\begin{enumerate} 
    \item We present \textit{the current state of evaluation methodologies} for ITSs by providing a comprehensive review of existing AIED evaluation research and pedagogical principles. This review covers a broad spectrum, ranging from evaluation approaches based on self-reports for traditional, static and rule-based ITSs, to NLG-based and pedagogically oriented evaluation for advanced deep learning and GenAI-based ITSs. 
    \item We pinpoint \textit{key challenges} in the evaluation of ITSs by highlighting the diverse pragmatics of evaluation and the lack of unification in the existing pedagogical principles, evaluation methodologies, and benchmarks. 
    \item Finally, we propose \textit{three feasible research directions} grounded in learning science principles and previous research in AIED, as a step towards mitigating the key challenges. These directions aim to build unified, scalable evaluation approaches that help track the progress and quality of GenAI-powered ITSs. 
\end{enumerate}

\section{Current State of Evaluation for ITS}
\vspace{-0.2cm}
\label{sec:background}
% Separate paragraphs for evaluation approaches across different time spans and types of evaluation, along with associated metrics.
% Add a figure to illustrate the progress of evaluation for ITS.

In this section, we present a comprehensive review of existing evaluation methodologies for ITSs, which broadly fall into three categories: {\em traditional}, {\em NLG-based}, and {\em pedagogy-oriented} evaluation methods.

\vspace{-0.3cm}
\subsection{Traditional Evaluation Approaches}
Traditional educational research on evaluating teacher effectiveness has a well-established and extensive history, as detailed by \cite{goe2008approaches} and \cite{muijs2006measuring}. This includes a range of methodological approaches for assessing teachers' practices, such as the analysis of classroom artifacts (e.g., evaluation of teacher assignments and student work), teaching portfolios, teacher self-reports of instructional practices via surveys, teaching logs, and interviews, as well as student evaluation of teacher performance. Pre-GenAI ITSs, including AutoTutor \cite{graesser2005autotutor}, DeepTutor \cite{rus2015deeptutor}, iDRIVE \cite{spreng2002idrive}, and Writing Pal (W-Pal) \cite{mcnamara2014writing} utilize rule-based and deep learning models to simulate human tutor behaviors, thereby supporting learning in areas such as STEM subjects, language acquisition, and computer literacy. The pedagogical effectiveness of these systems has been evaluated through role-simulation reports \cite{graesser2005autotutor} and empirical or extrinsic studies measuring learning gains \cite{nye2014autotutor}. For instance, previous research demonstrates that students interacting with AutoTutor show improvements in solving deep reasoning problems compared to those participating in traditional classroom-based learning \cite{nye2014autotutor}.

% adding AIED context
The AIED community are active contributors to emerging methodologies for measuring the effectiveness of human tutors and ITSs \cite{albacete2015dialogue,allen2015wrong,fang2024evaluating,schmucker2024ruffle}. For example, \cite{long2013skill} developed self-assessment for tutors in extrinsic studies based on self-regulatory learning theories. \cite{crossley2013using} developed an automated computational tool, Coh-Metrix, to evaluate the Writing Pal (W-Pal) ITS based on text cohesion. While these approaches have been effective in evaluating traditional human instructors and early AI-driven ITSs, their applicability to generative AI-powered tutoring systems raises significant concerns \cite{lin2023artificial}. Unlike their predecessors, generative AI models exhibit complex, context-dependent behaviors that pose unique challenges in evaluation. Existing methodologies, which often rely on static performance metrics or predefined criteria, may fail to capture the dynamic and adaptive nature of generative AI tutors \cite{macina-etal-2023-opportunities}. This gap highlights the need for new, more nuanced approaches that account for the variability and learning capabilities inherent in generative systems. Additionally, the reliance on traditional extrinsic frameworks may overlook critical aspects such as ethical considerations, bias in AI responses, and the alignment of AI-generated content with educational goals \cite{yan2024practical}. These concerns compel us to critically reassess the evaluation of GenAI-powered ITS effectiveness. %Our work sheds light on this challenging space and provides a learning sciences-based grounding for a path forward.
\vspace{-0.3cm}
\subsection{NLG-based Evaluation}
In recent years, general Natural Language Generation (NLG)-based domain-agnostic automated evaluation metrics have been leveraged to assess the quality of responses from ITSs \cite{jurenka2024towards}. These metrics rely on lexical and semantic matching to validate coherence, fluency, and human-likeness of the generated responses. For instance, \cite{macina2023mathdial} employed BLEU \cite{papineni2002bleu} and BERTScore \cite{zhang2019bertscore} to measure response quality of a generative AI-based tutor. The BEA 2023 shared task~\cite{tack-etal-2023-bea} incorporated additional metrics like DialogRPT \cite{gao2020dialogue} alongside BLEU and BERTScore to automatically evaluate ITS' responses developed by different teams. However, these metrics are not specifically designed to capture the pedagogical values inherent in effective tutoring \cite{tack2022ai}. Furthermore, these approaches require a gold reference to compute evaluation scores, which is often unavailable or non-unique \cite{al2024can}. Additionally, prior research has demonstrated the susceptibility of these models to superficial responses, where simple and generic outputs like ``Hello'' or ``teacher'' can misleadingly achieve high evaluation scores, thus undermining the robustness of these metrics~\cite{vasselli2023naisteacher}.
\vspace{-0.3cm}
\subsection{Pedagogically Oriented Evaluation}
As discussed in the previous two subsections, traditional evaluation approaches are primarily based on \textit{self-report} assessments, while NLG-based models often fall short in capturing pedagogical nuances. In this section, we review approaches that aim to capture the pedagogical values of GenAI-powered ITSs. The most reliable evaluation method remains human assessment; however, human evaluation is prone to subjectivity and personalized biases \cite{gehrmann2020repairing}. In the context of ITSs, additional challenges arise: the AIED and educational NLP communities accept multiple definitions of learning science principles \cite{jurenka2024towards,weinstein2018teaching}, leading to the absence of a unified and widely accepted protocol for pedagogical strategies and their associated definitions. For instance, \cite{tack2022ai} introduced an AI Tutor test considering only 3 pedagogical values, while \cite{koedinger2013instructional} presented 30 pedagogical strategies, and more recently, \cite{jurenka2024towards} proposed 28 distinct pedagogical strategies. Efforts have been made to unify these pedagogical frameworks, but they remain limited to specific tasks, such as single-turn mistake remediation \cite{maurya2024unifying}, although they establish a good footprint for future research. Automated evaluation metrics designed to capture pedagogical values often focus on a narrow subset of strategies, such as coherence in uptake \cite{demszky2021measuring}, or rely on training discriminator models \cite{macina2025mathtutorbench} to score pedagogical values; such approaches are frequently unreliable and incomplete. Specifically, \textit{meta-cognition} \cite{abdelshiheed2023leveraging,pinkwart2016another,porayska2016ai} and \textit{active learning} \cite{castro2021intelligent,rouzegar2024generative} have emerged as key dimensions for evaluating the performance of ITSs. However, both automated and human evaluation approaches currently lack reliability, scalability, and adaptability, making it difficult to consistently and efficiently assess the pedagogical effectiveness of these systems across diverse educational contexts. More recently, LLMs have been actively used as proxy judges \cite{gu2024survey}, though their evaluations remain largely restricted to metrics like helpfulness and human-likeness, often failing to capture the full spectrum of rich pedagogical values \cite{maurya2024unifying}. 

\section{Key Challenges in Evaluation of ITS}
\vspace{-0.2cm}
% List major challenges associated with evaluation.
The previous section outlined existing efforts and briefly touched upon the limitations of ITS evaluation. In this section, we will dive deeper and focus on two key challenges that hinder AIED progress in ITS evaluation methodologies. 
\vspace{-0.4cm}
\subsection{Challenge \#1: Diverse Pragmatics of Evaluation}

% \begin{itemize}
%     \item Participants type in benchmark creation: Expert or Novice
%     \item Participants type in benchmark creation: real learners, role-playing participants, or researchers
%     \item Benchmark type: single- or multi-turn
%     \item Benchmark type: unguided or scenario-guided
%     \item Evaluation type: human or automatic
%     \item Rater perspective: learners or educators
%     \item Evaluation scope: single turn or conversation level
%     \item Comparative evaluations: One-at-a-time or side-by-side
% \end{itemize}

One of the major challenges in evaluating GenAI-powered ITS lies in the \textit{diverse pragmatics of evaluation} \cite{jurenka2024towards}, which arise from the wide spectrum of design choices involved in benchmark creation and assessment methodologies. 

A fundamental consideration is the \textbf{type of participants} involved in benchmark creation \cite{brown2014make}. Depending on whether participants are experts, novices, real learners, role-playing participants, or researchers, the quality and nature of the benchmark data can vary significantly. Real learners bring authenticity to the evaluation process, reflecting genuine learning behaviors and challenges, while role-playing participants or researchers may offer more controlled and diverse scenarios, albeit with potential biases. Another crucial factor is the \textbf{structure of the evaluation benchmark} itself \cite{macina-etal-2023-opportunities,maurya2024unifying}. Benchmarks can be designed as single-turn or multi-turn interactions, each capturing different aspects of pedagogical effectiveness. Single-turn benchmarks focus on evaluating isolated responses, which can provide clarity but may overlook the dynamics of an evolving tutoring session. In contrast, multi-turn benchmarks capture the continuity and coherence of dialogue, which is essential to assess the system’s ability to adapt and respond to ongoing learner needs, but may overlook atomic utterance level responses. Moreover, the benchmarks creation can itself be \textbf{guided and unguided}. An unguided benchmark creation allows for open-ended interactions that better reflect natural conversations, while scenario-guided benchmarks offer structured contexts that help standardize evaluation and ensure coverage of specific learning objectives. This distinction affects the system’s ability to handle spontaneous and varied learner inputs as opposed to its proficiency in following defined pedagogical strategies.

As discussed in Section \ref{sec:background}, \textbf{evaluation methods} themselves differ significantly \cite{macina-etal-2023-opportunities}, ranging from human judgment to automated metrics. Human evaluation provides context-sensitive feedback but can be subjective and inconsistent, while automated evaluation offers scalability but often misses nuanced pedagogical values, requires good references and is not always reliable. Evaluation can be conducted at the single-turn level, focusing on immediate response quality, or at the conversation level, assessing the overall effectiveness and flow of the tutoring interaction. Conversation-level evaluation is particularly valuable for understanding long-term engagement and pedagogical impact. Finally, \textbf{comparative evaluation} introduces additional complexity, as systems can be evaluated one at a time, offering a focused assessment but risking inconsistent judgments, or side-by-side, allowing direct comparisons that highlight relative strengths and weaknesses. Each approach offers trade-offs between depth of evaluation and comparative clarity, underscoring the need for careful methodological choices when designing and interpreting ITS evaluation frameworks.
%Together, these diverse pragmatic considerations highlight the multifaceted nature of ITS evaluation, emphasizing the importance of aligning evaluation methodologies with the intended pedagogical goals and practical deployment scenarios.

\subsection{Challenge \#2: Lack of Evaluation Unification}
\label{sec:challenge2}

The implications of diverse pragmatics of evaluation lead to the \textit{absence of unification} across several critical dimensions. This lack of standardization impedes the systematic assessment and cross-comparison of different models, ultimately limiting progress in the development of effective and generalizable tutoring systems. We identify four primary categories contributing to this challenge.

\textbf{Lack of Unified Learning Science and Pedagogical Principles:} While different ITSs may have specialized goals and may justifiably focus on distinct learning science principles tailored to their specific use cases, there is a notable lack of uniformity even among systems designed for similar educational objectives \cite{jurenka2024towards}. Different systems leverage varying cognitive, constructivist, and behaviorist models \cite{anderson1995cognitive,piaget1959theory}, leading to inconsistencies in pedagogical strategies  \cite{vygotsky1978interaction}. As a result, this heterogeneity in both theoretical and pedagogical foundations hampers the reproducibility, scalability, and pedagogy-oriented evaluation of ITSs, ultimately limiting their effectiveness and comparability.

% \textbf{Lack of Unified Learning Science Principles:} Theoretical foundations of the learning science underpin the design and evaluation of ITS, yet there is no consensus on which principles to prioritize. Different systems leverage varying cognitive, constructivist, and behaviorist models \cite{anderson1995cognitive,piaget1959theory}, leading to inconsistencies in instructional strategies and knowledge representation. This heterogeneity hampers the reproducibility and scalability of findings across different ITSs.

% \textbf{Lack of Unified Pedagogical Values:} Effective tutoring requires alignment with pedagogical goals, such as fostering conceptual understanding, promoting critical thinking, and encouraging metacognitive skills \cite{vygotsky1978interaction}. However, ITS implementations often reflect divergent pedagogical priorities, with some focusing on procedural fluency and others on exploratory learning. This misalignment complicates the development of pedagogy-oriented evaluation systems. 

\textbf{Lack of Unified Evaluation Metrics:} As noted in Section \ref{sec:background}, ITS evaluation uses automated methods, which are scalable but often unreliable, and human evaluation, which is reliable but costly and non-scalable \cite{chi2008active}. This dichotomy results in ambiguous interpretations of ITSs performance and undermines the reliability of comparative analyses. Standardized metrics are crucial for establishing clear benchmarks and tracking incremental improvements. Efforts of \cite{maurya2024unifying} provide a foundation and promise for unification in this direction.

\textbf{Lack of Unified Evaluation Benchmarks:} Benchmark datasets are essential for the validation and generalization assessment of models, particularly in ITSs. However, there is a notable dearth of comprehensive and diverse datasets specifically designed for ITS applications, which significantly hinders progress in this area \cite{macina-etal-2023-opportunities}. The current benchmarks often fall short in terms of domain coverage and interaction depth, limiting their ability to accurately capture the intricate dynamics of real-world educational dialogues. This gap impedes the development of pedagogically sound, robust, and scalable ITS design. Table \ref{tab:benchmarkes} provides an overview of some of the popular datasets used in ITS evaluation, highlighting their respective domains, dialogue characteristics, and grounding mechanisms. Despite their utility, these datasets fail to provide the necessary breadth and complexity required for a holistic assessment of ITS.

\begin{table}[ht]
\vspace{-0.5cm}
\centering
\resizebox{\textwidth}{!}{%
\begin{tabular}{|l|c|l|r|r|l|l|}
\hline
\textbf{Dataset} & \textbf{Synthetic} & \textbf{Domain} & \textbf{\# Dialogues} & \textbf{\# Moves} & \textbf{Grounding} & \textbf{Setting} \\
\hline
CIMA  \cite{stasaski2020cima}    & No     & Language  & 391   & max 5    & Image, answer        & 1:1       \\
Bridge  \cite{wang-etal-2024-bridging}  & No     & Math      & 700   & max 10   & Image, confusion     & 1:1       \\
MathDial \cite{macina2023mathdial} & Mixed    & Math      & 2861  & max 4    & Confusion, answer    & 1:1       \\
TSCC  \cite{caines2020teacher}   & No     & Language  & 102   & max 5    & None                 & 1:1       \\
TalkMoves \cite{suresh2022talkmoves} & No     & Science   & 567   & max 10   & None                 & Classroom \\
NCTE  \cite{demszky2022ncte}    & No     & Math      & 1660  & --   & None                 & Classroom \\
MRBench \cite{maurya2024unifying}  & Mixed  & Math      & 200   & max  10   & Confusion, answer    & 1:1       \\
\hline
\end{tabular}%
}
\vspace{0.1cm}
\caption{Overview of benchmarks used in ITS evaluation}
\vspace{-0.5cm}
\label{tab:benchmarkes}
\end{table}

\section{Path Forward}
\vspace{-0.3cm}
In the previous two sections, (1) we reviewed the current state of evaluation from different perspectives and (2) pinpointed the key challenges that hinder the progress of evaluation methodologies for ITSs. These highlight multiple research opportunities that help bridge the ITS evaluation gap. In this section, we take a \textit{step forward} and propose three key research directions that can help mitigate the existing challenges and provide a framework for the holistic evaluation of ITSs. These directions are inspired by the gaps observed in the ITS evaluation literature and are grounded in previous research in AIED  and learning science principles. However, we acknowledge that this is just a preliminary effort toward addressing numerous open problems, and we call for the AIED research community to join hands in this endeavor. As discussed in Section \ref{sec:background}, \textit{meta-cognition} and \textit{active learning} are well-established key principles of ITS design. Building on these foundations, this proposal aims to develop automated, scalable, and reliable evaluation methodologies to assess these two critical dimensions. Although the proposed research directions are targeted for the evaluation of GenAI-powered ITSs, it is also applicable to human tutors.

\vspace{-0.3cm}
\subsection{Research Line \#1: Evaluation Unification}
\vspace{-0.2cm}
% Introduction
% Example/Figure (optional)
% Research Statement
% Potential Approach
% (a) Grounding to previous AIED research
% (b) Connection to Learning Science Principle 

The challenges mentioned in Section \ref{sec:challenge2} highlight that the lack of a unified evaluation framework is one of the key obstacles to the progress in ITS development. While we acknowledge that a unified framework may introduce social biases related to different learning goals, subjects, and contexts within ITSs, we still believe that establishing a standardized approach led by the AIED community would provide a strong foundation for future research and facilitate easier adaptability. As an initial step, for focused educational use cases such as student mistake remediation and feedback generation, a unified framework should be developed based on existing work. This framework should incorporate key learning science principles, effective pedagogical strategies, standardized evaluation metrics, and well-defined benchmarks. 

\textbf{Research Statement:} Current evaluation practices for ITSs often focus on isolated aspects of system performance, resulting in fragmented insights. We hypothesize that a unified evaluation framework, bringing together AIED researchers to develop widely accepted evaluation protocols and benchmarks for similar use cases, will enable a more holistic assessment of ITSs.

\textbf{Potential Approach:} Let’s consider a case study from the recent work by \cite{maurya2024unifying} to convince readers and assess the feasibility of this research direction. In \cite{maurya2024unifying}, the authors proposed a unified evaluation taxonomy to systematically assess the ITSs' and human tutor's current response quality given the conversation history for the student mistake remediation (SMR) task. They extensively reviewed the existing literature and concluded that several existing works only partially address the essential pedagogical properties required for effective tutoring. Based on this analysis, they proposed an eight-dimensional evaluation taxonomy grounded in established learning science principles. Furthermore, they annotated and released the \texttt{MRBench} benchmark, developed based on this taxonomy, to quantitatively measure the pedagogical effectiveness of ITSs. This effort, though focused on the SMR task in educational settings, presents a strong proof of concept for the unification of evaluation frameworks. Future research can develop automated evaluation metrics for each dimension to enhance scalability and reliability. This effort lays a foundation for future work toward a unified evaluation paradigm, which should include a\textit{ comprehensive review of the existing literature}, \textit{high-quality dataset creation}, the \textit{formulation of an evaluation taxonomy} aligned with pedagogical principles, \textit{rigorous training of annotators}, \textit{detailed annotation processes}, and the \textit{development of robust automated evaluation metrics} for large-scale deployment.

\vspace{-0.3cm}
\subsection{Research Line \#2: Measuring Pedagogical Guidance}
\vspace{-0.3cm}
% Introduction
% Example/Figure (optional)
% Research Statement
% Potential Approach
% (a) Grounding to previous AIED research
% (b) Connection to Learning Science Principle 

For any learner, successful learning enables the generalization of skills beyond the given context and fosters deep, reflective thinking. To cultivate these skills, tutors often provide pedagogical guidance in the form of hints, examples, explanations, acknowledgments, and constructive feedback. This approach is rooted in the principle of metacognition \cite{abdelshiheed2023leveraging,dehaene2021we} from learning sciences. By offering timely and context-aware support, tutors help students navigate their learning journey by promoting mistake discovery and providing constructive feedback (both positive and negative), among others.

\textbf{Research Statement:} This research direction aims to develop a robust quantitative framework to assess the appropriateness and richness of the pedagogical guidance provided by GenAI-powered ITSs. Specifically, given a partial conversation between a student and a human tutor or ITS, we propose methods to systematically evaluate the effectiveness of the tutor’s next response in supporting the student’s metacognitive development. %This will be achieved by measuring the \textit{pedagogical appropriateness}, \textit{contextual relevance}, and \textit{factual correctness} offered in the tutor's response.

\begin{figure}[ht]
\vspace{-0.8cm}
    \centering
    \begin{minipage}{0.5\textwidth}
        \centering
        \includegraphics[width=\linewidth]{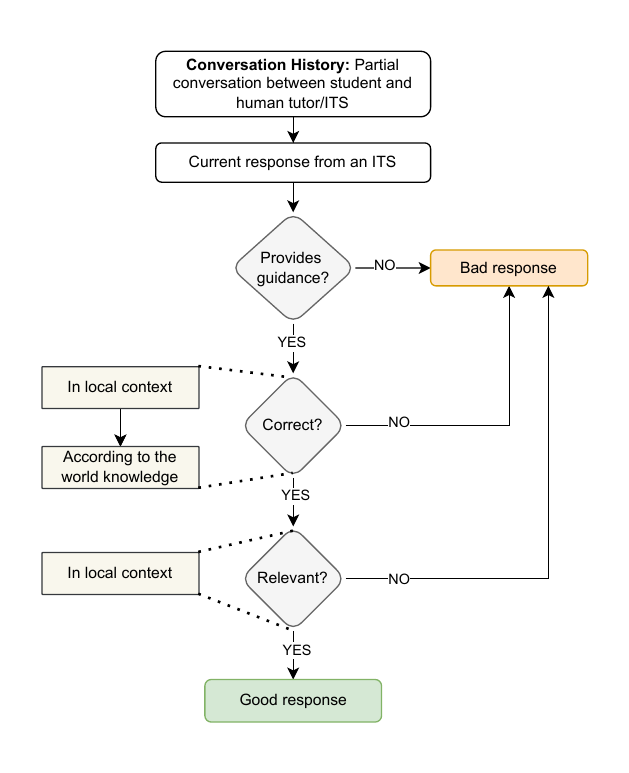}
        % \caption{Caption for Figure 1}
        % \label{fig:figure1}
    \end{minipage}\hfill
    \begin{minipage}{0.5\textwidth}
        \centering
        \includegraphics[width=0.6\linewidth]{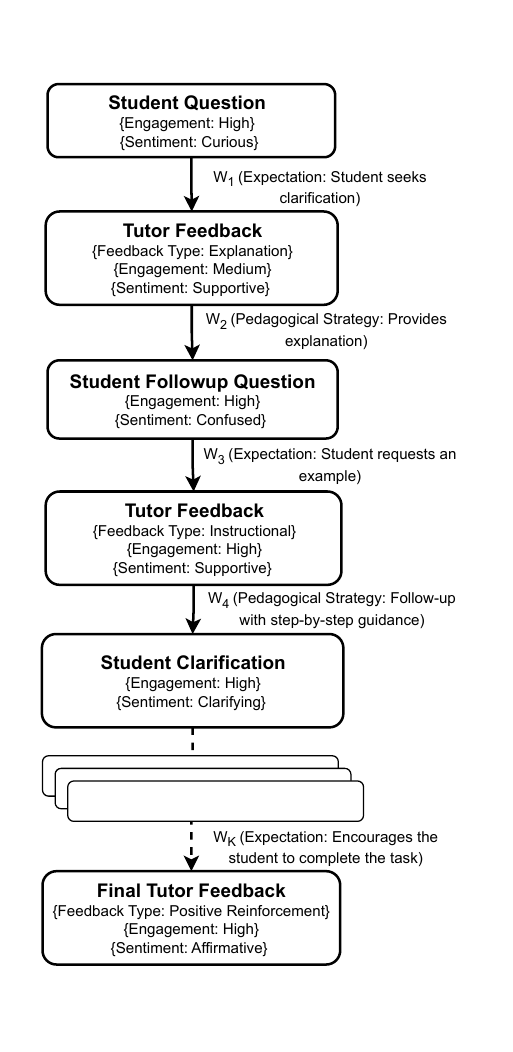}
        % \caption{Caption for Figure 2}
        
        \label{fig:figure2}
    \end{minipage}
    \vspace{-0.5cm}
    \caption{(Left) Flowchart illustrating good and bad tutor responses to measure pedagogical guidance. (Right) Conversation flow graph to measure the student's active learning, where nodes represent different utterance features and edge weights indicate the intensity of conversational dependency.}
    \label{fig:peda_guide}
    \vspace{-1.cm}
\end{figure}

\textbf{Potential Approach:} In existing AIED research \cite{abdelshiheed2023leveraging,pinkwart2016another,porayska2016ai}, metacognition is one of the most frequently cited  capabilities of ITSs. However, there is a lack of scalable and automated methodologies to assess how well a tutor's guidance aligns pedagogically with metacognitive principles. Building on prior research \cite{jurenka2024towards,khuwaja1996intelligent}, we consider a tutor's next response appropriate if it provides \textit{pedagogical guidance}, is \textit{factually accurate}, and remains \textit{contextually relevant}. See Figure \ref{fig:peda_guide} (left) for a visual illustration of these principles.

A modular approach to modeling each of the three aforementioned properties offers a promising research direction. These modules can collectively help to develop a holistic discriminator reward model to distinguish between \textit{good} and \textit{bad} responses. \textit{Pedagogical guidance} can be approached  as identifying the student’s current state (e.g., a confusion) and selecting the most appropriate pedagogical strategy (e.g., a hint, examples, etc.) based on that state. This can be achieved by leveraging prior work \cite{wang-etal-2024-bridging,wang2024tutor}, where human tutor models and language model-based classifiers were employed. For \textit{contextual relevance}, sentiment embedding-based approaches can be adapted as proposed by \cite{demszky2021measuring,zhang2019bertscore}. Finally, the most challenging aspect is the identification of \textit{factual correctness}, which can apply at the local level (where the current response contradicts statements in the conversation history) or at the global level (inconsistencies with established factual knowledge). This issue is particularly prevalent in GenAI-powered ITSs, as they are prone to hallucinations \cite{huang2025survey}. At the local level, this can be framed as a textual entailment problem, while at the global level, prior research on factuality verification \cite{wang2024factuality} offers valuable insights and a path forward.   

% \begin{figure}
%     \centering
%     \vspace{-0.8cm}
%     \includegraphics[width=0.5\linewidth]{measure_pedagogy_v2.pdf}
%     \vspace{-0.5cm}
%     \caption{Flowchart to illustrate good and bad tutor responses based on pedagogical principles.}
%     \vspace{-1cm}
%     \label{fig:peda_guide}
% \end{figure}

\vspace{-0.3cm}
\subsection{Research Line \#3: Measuring Active Learning}
Active learning is a critical component of effective tutoring, allowing students to engage actively in the learning process through critical thinking and reflection. This approach aligns with well-established learning science principles, such as constructivism and inquiry-based learning \cite{piaget1972theory,vygotsky1978mind}. Effective tutors encourage active engagement by posing thought-provoking questions, guiding students without offering immediate answers, and maintaining open-ended conversations to promote reflective thinking.

\textbf{Research Statement:} This research direction should aims to develop a systematic framework for evaluating and measuring the active learning capabilities of ITSs. Specifically, for an ongoing dialogue between a student and an ITS, we propose methods to assess how effectively the tutor stimulates \textit{active engagement}, \textit{adapts to the student's emotional and cognitive states}, and provides \textit{encouraging feedback}.

\textbf{Potential Approach:} Previous research in AIED \cite{castro2021intelligent,rouzegar2024generative} has emphasized the importance of active learning strategies. However, scalable and automated methods to measure these strategies remain underexplored. We believe that a graph-based modeling approach is a promising direction for assessing the active learning capabilities of tutoring systems. Within this approach, the dialogue between the student and tutor is represented as a dynamic, directed \textit{conversation flow graph} \cite{scarselli2008graph,wu2020comprehensive}, where each node represents an individual utterance, and edges capture conversational dependencies, such as follow-up questions, clarifications, or feedback. Figure \ref{fig:peda_guide} (right) presents a sketch of the flow graph. Node features encode engagement signals, sentiment, and feedback types, while edge weights $W_i$ represent the strength of interactions.

The conversation flow graph can be modeled using a graph neural network (GNN) \cite{kipf2016semi,velivckovic2017graph}, enabling the prediction of engagement, adaptive support, and encouraging feedback from its structure. By capturing both content and conversational dynamics, the model can identify student engagement patterns like sustained reflective dialogue and effective adaptation to student states. This approach allows for a more nuanced analysis of the tutoring process, distinguishing between productive and unproductive interactions. Furthermore, the GNN framework can generalize across diverse ITS datasets, making it a scalable solution for evaluating different tutoring strategies. Additionally, attention mechanisms in GNNs \cite{wu2020comprehensive} can highlight influential interactions, offering insights into tutor behavior that drive active learning and promote student understanding.

Other modeling approaches beyond GNNs can be used to represent the conversation flow graph. For instance, Graph Attention Networks (GATs) \cite{velivckovic2017graph} can be employed to dynamically assign importance weights to neighboring nodes, allowing the model to focus on critical conversational turns such as key feedback or pivotal clarifications. Additionally, Graph Transformers \cite{yun2019graphtransformer} offer a promising alternative by leveraging global attention mechanisms to capture long-range dependencies across the dialogue, thus enabling a more holistic understanding of complex tutoring interactions. These advanced architectures could further enhance the model’s ability to identify nuanced patterns of student engagement and tutor support strategies.

% Introduction
% Example/Figure (optional)
% Research Statement
% Potential Approach
% (a) Grounding to previous AIED research
% (b) Connection to Learning Science Principle 

% \section{Implications of Research Directions}
% Details of associated implications of the stated research directions, if any.
\vspace{-0.3cm}
\subsection{Implications of Proposed Research Directions}
\vspace{-0.3cm}
If developed and deployed, these research directions could significantly enhance ITSs. A \textit{unified evaluation framework} would standardize assessments, promoting consistency and transparency across systems. While it may oversimplify the diverse needs in different educational contexts, careful design can mitigate the risk of bias and ensure that specific learning goals are not overlooked.
Furthermore, taxonomy-driven evaluation serves as a foundational step toward systematically assessing ITSs. At the same time, it offers limited reflection of real-world tutoring quality, where student interactions are often complex and non-linear. The next step should involve conducting real-world randomized controlled experiments to rigorously evaluate the effectiveness of any tutoring system. Finally, unification efforts must strike a balance: evaluation should not be overly fine-grained—such as requiring a separate taxonomy for each utterance—nor overly coarse, where a tutor is assumed to be holistically optimal. This level of granularity in evaluation design is critical for the effectiveness of taxonomy- and benchmark-driven research in ITSs. \textit{Measurement of pedagogical guidance} would improve the adaptability of ITS, ensuring that guidance aligns with students' metacognitive needs, fostering deeper engagement. Although assessing the subtleties of student cognition and emotions is challenging, thoughtful integration of adaptive mechanisms could enhance the accuracy and appropriateness of interventions. \textit{Measuring active learning} would allow systems to better stimulate critical thinking, adapt to student states, and provide timely, context-sensitive feedback, enhancing overall learning experiences. While graph-based models may face scalability issues and complexity in capturing dynamic interactions, ongoing advancements in GenAI-based computational models can address these challenges. Together, these advancements could lead to more effective, scalable, and personalized ITSs, transforming educational practices.

\section{Conclusion}
\vspace{-0.2cm}
% Summarizing the key takeaways.
The remarkable performance of GenAI models has led to active development and new opportunities for dialogue-based ITSs in the AIED field. However, the true progress of ITSs remains unclear due to the lack of reliable and robust evaluation measures. This challenge arises from the diverse pragmatics of evaluation and the non-uniform adoption of subjective and personalized protocols and benchmarks. This work highlights this gap by presenting the current state of ITS evaluation from multiple perspectives, identifying key challenges, and proposing three potential and feasible research directions rooted in pedagogical principles and existing AIED research. These directions pave the way towards the establishment of reliable, scalable, and robust evaluation frameworks for ITSs. We believe that this work provides a comprehensive overview of the existing landscape and offers valuable insights to advance the evaluation of ITSs.

\section*{Acknowledgments}
\label{sec:ack}
We are grateful to Google for supporting this research through the Google Academic Research Award (GARA).

\vspace{-0.3cm}
\bibliographystyle{splncs04}
\bibliography{bibliography}
\vspace{-0.4cm}

% \bibitem{hursh2007assessing} Hursh, D. (2007). Assessing No Child Left Behind and the rise of neoliberal education policies. \textit{American Educational Research Journal}, 44(3), 493–518.

% \bibitem{au2007high} Au, W. (2007). High-stakes testing and curricular control: A qualitative metasynthesis. \textit{Educational Researcher}, 36(5), 258–267.

% \bibitem{hargreaves2003teaching} Hargreaves, A. (2003). \textit{Teaching in the knowledge society: Education in the age of insecurity}. Teachers College Press.

% \bibitem{perelman2014critiquing} Perelman, L. (2014). Critiquing automated essay scoring systems: A challenge for educational assessment. \textit{Journal of Writing Assessment}, 7(1), 1–15.

% \bibitem{haswell2006automated} Haswell, R. (2006). Automated essay scoring and the future of writing assessment. \textit{College Composition and Communication}, 58(1), 36–63.

% \bibitem{perelman2012construct} Perelman, L. (2012). Construct validity, length, and automated essay scoring. \textit{Educational Leadership}, 70(4), 44–49.

% \bibitem{whitlock2020limitations} Whitlock, A. (2020). The limitations of automated feedback in educational assessment. \textit{Educational Technology Review}, 18(2), 122–137.

% \end{thebibliography}
\end{document}